%% file: Paper.tex
\documentclass[conference]{IEEEtran}
\IEEEoverridecommandlockouts
\usepackage[dvipsnames]{xcolor}

\definecolor{mydarkblue}{rgb}{0,0.08,0.45}
\definecolor{myblue}{HTML}{3b75c3}
\definecolor{myred}{HTML}{E33222}
\definecolor{mygreen}{HTML}{438773}
\definecolor{mymaroon}{RGB}{142,27,19}
\definecolor{maroon}{HTML}{800000}
\definecolor{mycite}{cmyk}{0.55,1,0,0.15}
\definecolor{codeblue}{rgb}{0.25,0.5,0.5}
\definecolor{codekw}{rgb}{0.85, 0.18, 0.50}
\definecolor{codegreen}{rgb}{0,0.6,0}
\definecolor{codegray}{rgb}{0.5,0.5,0.5}
\definecolor{codepurple}{rgb}{0.58,0,0.82}
\definecolor{backcolour}{rgb}{0.95,0.95,0.92}
\definecolor{mygray}{gray}{0.925}

\usepackage{cite}
\usepackage{amssymb}
\usepackage{textcomp}

\usepackage{adjustbox}
\usepackage{algpseudocode}
\usepackage{amsfonts}
\usepackage{amsmath}
\usepackage{array}
\usepackage{bbm}
\usepackage{bm}
\usepackage{booktabs}
\usepackage{caption}
\usepackage{color}
\usepackage{colortbl}
\usepackage{csquotes}
\usepackage{enumitem,makecell}
\usepackage{epsfig}
\usepackage{etoc}
\usepackage{float}
\usepackage{graphicx}
\usepackage[group-separator={,},group-minimum-digits=4]{siunitx}
\usepackage{hyperref}     
\usepackage[capitalize,noabbrev,nameinlink]{cleveref}
\usepackage{latexsym}
\usepackage[linesnumbered,ruled]{algorithm2e}
\usepackage{lipsum}
\usepackage{titlesec}
\usepackage{listings}
\usepackage{makecell}
\usepackage{mathtools}
\usepackage{multirow}
\usepackage{pgfgantt}
\usepackage{pifont}
\usepackage{rotating}
\usepackage{soul}
\usepackage{subcaption}
\usepackage{url,afterpage,amsfonts}
\usepackage{verbatim}
\usepackage{wrapfig}
\usepackage{xcolor}
\usepackage{xspace}
\usepackage{array,xcolor,colortbl}
\usepackage{multirow}
\def\BibTeX{{\rm B\kern-.05em{\sc i\kern-.025em b}\kern-.08em
    T\kern-.1667em\lower.7ex\hbox{E}\kern-.125emX}}

\newcounter{magicrownumbers}

\hypersetup{
    colorlinks=true,
    citecolor=mycite,
    urlcolor=mydarkblue,
    linkcolor = mydarkblue
}

\begin{document}

\title{Entropy Aware Training for Fast and Accurate Distributed GNN}

\author{\IEEEauthorblockN{Dhruv Deshmukh\IEEEauthorrefmark{1},
Gagan Raj Gupta\IEEEauthorrefmark{2}, Manisha Chawla\IEEEauthorrefmark{3}, Vishwesh Jatala\IEEEauthorrefmark{4} and
Anirban Haldar\IEEEauthorrefmark{5}}
\IEEEauthorblockA{Department of CSE,
IIT Bhilai, India\\
Email: \{\IEEEauthorrefmark{1}dhruvr,
\IEEEauthorrefmark{2}gagan,
\IEEEauthorrefmark{3}manishach,
\IEEEauthorrefmark{4}vishwesh,
\IEEEauthorrefmark{5}anirbanh\}@iitbhilai.ac.in
}
}

\maketitle

\input{Sections/0_Abstract}


\input{Sections/1_Introduction}
\input{Sections/2_RelatedWork}
\input{Sections/3_Approach}
\input{Sections/4_ExperimentalSetup}

\input{Sections/5_Results}

\input{Sections/6_Conclusion}

\section*{Acknowledgment}
The support and the resources provided by PARAM Shakti at the Indian Institute of Technology, Kharagpur and PARAM Sanganak at the Indian Institute of Technology, Kanpur under the National Supercomputing Mission, Government of India are gratefully acknowledged. Vishwesh Jatala acknowledges the funding received from DST/SERB through Start-up Research Grant SRG/2021/001134.

\bibliographystyle{IEEEtran}
\bibliography{References/datasets,References/GNN,References/DistGNN,References/GNNApplications,References/GPUGNN,References/misc,References/GraphAnalytics/graphs,References/GraphAnalytics/gpugraphs,References/GraphAnalytics/outofcore,References/GraphAnalytics/partitioning,References/GraphAnalytics/resilience,References/GraphAnalytics/others}

\input{Sections/7_Appendix}

\end{document}

%% file: Sections/0_Abstract.tex
\begin{abstract} \label{sec:abstract}
Several distributed frameworks have been developed to scale Graph Neural Networks (GNNs) on billion-size graphs. On several benchmarks, we observe that the graph partitions generated by these frameworks have heterogeneous data distributions and class imbalance, affecting convergence, and resulting in lower performance than centralized implementations. We holistically address these challenges and develop techniques that reduce training time and improve accuracy. We develop an Edge-Weighted partitioning technique to improve the micro average F1 score (accuracy) by minimizing the total entropy. Furthermore, we add an asynchronous personalization phase that adapts each compute-host's model to its local data distribution. We design a class-balanced sampler that considerably speeds up convergence. 
We implemented\footnote{Code available at \url{https://github.com/Anirban600/EAT-DistGNN}} our algorithms on the DistDGL framework and observed that our training techniques scale much better than the existing training approach. We achieved a (2-3x) speedup in training time and 4\% improvement on average in micro-F1 scores on 5 large graph benchmarks compared to the standard baselines. 
\end{abstract}

%% file: Sections/1_Introduction.tex
\section{Introduction} \label{sec:intro}

Graph neural networks (GNNs) have made tremendous progress in recent years and have achieved state-of-the-art performance in diverse applications~\cite{largescalesurvey}, including social network analysis, credit-card fraud detection~\cite{PickNChoose}, recommender systems, etc.
A core operation in GNNs is message-passing to aggregate information from neighbors, which is then used to learn task-specific vertex embeddings. These operations require internal graph structures to be stored (in memory) during forward and backward propagation, making it very difficult to scale on industrial-grade graphs with billion-scale edges. Several distributed GNN frameworks~\cite{DistDGL,P3} have been developed recently to address this problem. 

One of the critical steps in most of the above frameworks~\cite{DistDGL} to achieve scalability is partitioning the input graph into disjoint sub-graphs of similar size and assigning one sub-graph (partition) to every compute host in the cluster for training. During the distributed training process, in every iteration, a sample (batch) of labeled training nodes is used by every compute host to estimate the local gradients, using which a global average is computed and used to update the global model. The final model obtained at the end of training performs tasks such as label prediction of the test vertices present at each compute host in parallel. 

\begin{figure}
    \centering
    \includegraphics[scale=0.35]{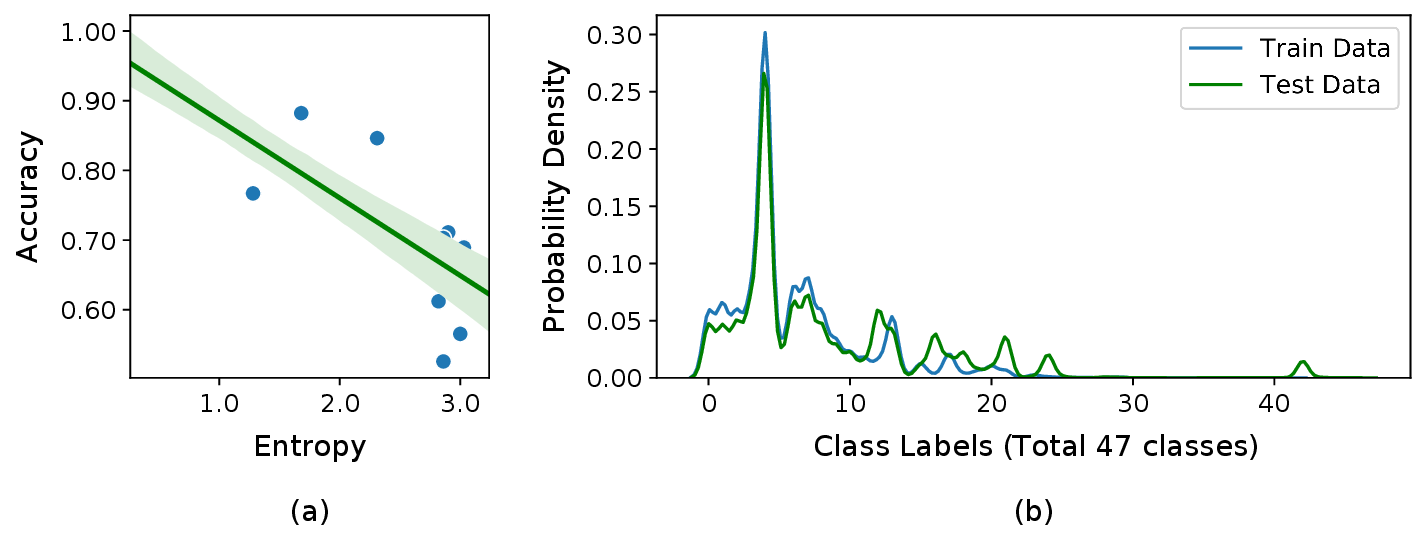}
    \vskip -3mm
    \caption{Challenges in Distributed GNN Training. a) Variation in the entropy of partitions affects accuracy in OGBN-Products dataset with 16 partitions. A linear regression line with 95\% confidence interval is overlayed.
    b) Class imbalance and out-of-distribution in OGBN-Products. 
    }
    \label{fig:EntropyPapers}
    \vskip -4mm
\end{figure}

Similar to \cite{cluster-gcn}, we observe in Fig.~\ref{fig:EntropyPapers}a), that the entropy\footnote{ Entropy = $-\Sigma_i p_ilog(p_i)$ where $p_i$ is the fraction of the i-th label} of these partitions has significant variation due to non-i.i.d labels. This affects convergence results in lower performance than centralized implementations, especially when number of compute hosts is increased. Fig.~\ref{fig:EntropyPapers}a) shows the micro average F1 scores on the test set per compute node after distributed training on the OGBN-Products dataset with 16 compute hosts. We observe that the compute hosts having partitions with lower entropies typically achieve higher accuracy (same as micro avg. F1 score), although there are some outliers due to complex factors involved. Fig.~\ref{fig:EntropyPapers}b) shows the huge imbalance in the label distribution in the OGBN-Products dataset, which leads to a bias in the models toward majority classes. Graph partitioning can worsen this further.

We holistically address these challenges and develop techniques that reduce training time and improve accuracy. We extensively study the impact of the entropy of partitions on the performance of GNNs. We design an efficient edge-weighted partitioning algorithm that aims to minimize partitions' total entropy to improve micro average F1 scores. In this algorithm, the weight of the edge is assigned based on the node degree and similarity of features of the corresponding adjacent nodes. Since nodes with similar labels often have similar features, their edge weights are higher. Edge-weighted partitioning schemes aim to minimize the sum of edge weights of cut edges, resulting in partitions with nodes of similar labels and thus reduce total entropy. 

We then investigate personalization \cite{Ditto, AdaGNN} of models at each compute node in an asynchronous mode to adapt to the local data distributions. 
This results in local personalized models at each compute node, significantly improving performance. At the same time, this helps in faster training as the communication and synchronization costs of averaging gradients across all partitions are reduced greatly.

Finally, we develop a class-balanced sampler (CBS), that promotes equitable representation of minority classes in every batch during training. 
For large graphs, our sampler reduces the number of training examples (from the majority classes) per epoch, resulting in a faster completion time per epoch. We implement these techniques in the DistDGL framework~\cite{DistDGL} and perform extensive experiments on large-scale graph benchmarks.  We briefly summarize the contributions of our paper:
\begin{itemize}
    \item Development of novel entropy-aware partitioning algorithms to minimize the total entropy 
    \item Development of a new class-balanced sampler to address the class-imbalance problem in real-world graphs and speed-up training.
    \item Asynchronous personalization of graph models to local data distributions to achieve higher performance.
    \item Extensive experimental evaluation on large-graph benchmarks including OGBNPapers-100M with a billion edges on commodity HPC clusters. Our techniques achieve a speedup of 2-3x, improve micro-F1 score by 4\% and 
    on average in 5 large graph benchmarks compared to the standard DistDGL implementation.
\end{itemize}

The rest of the paper is organized as follows. We provide the relevant background for distributed GNN and related research in Section~\ref{sec:related}. This is followed by a detailed description of our algorithms for entropy aware distributed GNN training in Section~\ref{sec:approach}. We then describe the datasets and experimental setup in Section~\ref{sec:experiments}. The experimental results and ablation studies are discussed in Section~\ref{sec:results} followed by the conclusions.

%% file: Sections/2_RelatedWork.tex
\section{Background and Related Work}\label{sec:related}
Graph neural networks learn representations of vertices/edges by aggregating the messages received from the neighborhood vertices. Consider a graph $G(V,E)$ having $V$ vertices, $E$ edges, $L$ Labels. Each vertex (or edge) in the graph is associated with a feature vector, $\mathbf{x_v}$, which is used to initialize $\mathbf{h_v}^0$. For a GNN consisting of $k$ layers ($k>0$), the embedding $\mathbf{h}^i_v$ of $v$ at the $i^{th}$ layer is computed from the embeddings generated by the previous layer by applying an aggregation function $AGG_i$ and a non-linear function $\sigma_i$ on the messages received from the neighborhood of $v$ (denoted as $N(v)$), its own feature after multiplying with the weight matrix $\mathbf{W}^i$. The equations to compute $\mathbf{h}^i$ from $\mathbf{h}^{i-1}$ are listed below. 
\begin{equation} \label{eq:aggregation}
\mathbf{h}^i_{N(v)} = AGG_i(\mathbf{h}^{i-1}_{u}, \forall{u\in N(v)})
\end{equation}
\vskip -3mm
\begin{equation}  \label{eq:outputfeature}
\mathbf{h}^i_{v} = \sigma_i(\mathbf{W}^i.CONCAT(\mathbf{h}^i_{N(v)}, \mathbf{h}^{i-1}_{v}))
\end{equation}

The final embeddings, $\mathbf{h}^k_v$, are used to make the predictions, and the gradients of the loss w.r.t. $\mathbf{W}^i$ are used to update $\mathbf{W}^i$ till convergence is reached.

Many GNN models were developed to improve accuracy; these models differ in the way the aggregations are performed. 
GraphSAGE~\cite{graphsage} aggregates neighborhood vertices features along with its own features. It has been widely used for distributed settings as it uses neighborhood sampling techniques to reduce computation and communication overhead. It has been observed that GraphSAGE achieves reasonable performance when trained with 2-layer GNN, and adding further layers causes over-fitting and increases communication overhead~\cite{En-GCN, Graph-Gym}. In this paper, we have used the GraphSAGE algorithm in all the experiments. 


\begin{figure*}[ht] 
        \centering{
		\includegraphics[scale=0.25]{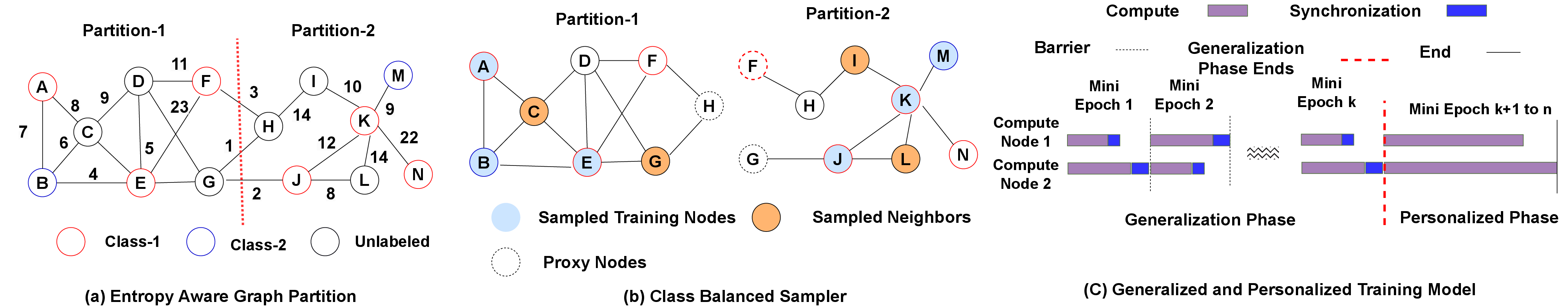}
        }
	\caption{Overview of the proposed approaches for high-performance distributed GNN.}
	\label{fig:overview}
	\vskip -6mm
\end{figure*}


\textbf{Distributed GNN Frameworks:} DistDGL~\cite{DistDGL}, PipeGCN~\cite{PipeGCN}, and Adgraph~\cite{ADGRAPH} adopt METIS graph partitioning. P3~\cite{P3} uses random hash, and DistGNN~\cite{DistGNN} uses vertex cut partition to partition the graph. However, these frameworks do not consider class imbalance.

 
DistDGL is based on distributed PyTorch and has configurable functions for partitioning, sampling, etc., which we use to implement our algorithms. Our algorithms proposed in this paper focus on improving various performance metrics while reducing training time in the distributed setting.


\textbf{Graph Partitioning:} Distributed graph analytics, or GNN requires partitioning the input graph among multiple hosts. Graph partitioning has been studied for decades~\cite{metispowerlaw,stanton12,metis}. METIS~\cite{metis} has been widely used to partition a graph in distributed GNN models~\cite{DistDGL,AliGraph}. It minimizes the number of edge cuts among the partitions and balances the nodes in each partition. Most partition techniques achieve two primary goals: (1) balance computation load among the hosts and (2) reduce communication overhead among the hosts. 
To the best of our knowledge, none of the partitioning techniques have considered minimization or balancing entropy to improve distributed GNNs' performance.

\textbf{Graph Sampling:} It has been observed that GNN-based algorithms have poor performance when the label distribution is skewed. To address this problem, \cite{PickNChoose, GraphSMOTE, GraphSAINT} were developed. GraphSMOTE \cite{GraphSMOTE} attempts to extend the classical approach of synthetic minority oversampling algorithms to GNNs by sampling in the representation space and using an edge generator. 
PC-GNN \cite{PickNChoose} uses a label-balanced sampler to construct sub-graphs and neighbors for message aggregation during training for binary classification problems. 
We generalize these techniques to multi-class node classification and distributed GNN training. 

\textbf{Personalization:} Personalization or fine-tuning of models is an important step in dealing with the heterogeneity of clients. This has been studied recently in the Federated learning literature~\cite{personalized, splitgp, AdaGNN, Ditto} but hasn't been evaluated thoroughly in the context of distributed GNNs. Personalization improves accuracy when the test data distribution matches with the train data distribution. Ada-GNN \cite{AdaGNN} used a basic form of personalization in a centralized implementation. 

\textbf{Performance Metrics: }
Micro-F1 computes accuracy using all testing examples at once, while Weighted Macro-F1 (Weighted-F1) is computed by giving a weight proportional to the class frequency during the averaging of individual F1 scores.
The effectiveness of the distributed GNN is also measured using two other metrics: training time and epoch time. Training time denotes the maximum training time across the compute hosts. Epoch time denotes the time taken for one complete pass through the entire training dataset.



%% file: Sections/3_Approach.tex

\section{Our Algorithms} \label{sec:approach}

Figure~\ref{fig:overview} gives an overview of the proposed approaches. First, we construct a weighted graph from the unweighted graph, where the weight of an incoming edge is determined by the node degree and feature similarity of the adjacent nodes. Then, we partition the weighted graph among the compute hosts using the weighted METIS~\cite{metis}, as shown in Figure~\ref{fig:overview}(a). 

Once the graph is partitioned, we use a class-balanced sampler (CBS) to select a subset of training nodes during a mini-epoch, as shown in Figure~\ref{fig:overview}(b). CBS samples training nodes of the minority classes with higher probability and nodes of majority classes with lower probability. Once the mini-epoch training nodes are chosen, a random training batch is created during each iteration. K neighbors are sampled for each node in this batch to perform message aggregation using the update Equation (\ref{eq:aggregation}). 

The distributed training occurs in two phases: generalization followed by personalization, as shown in Figure~\ref{fig:overview}(c). The training starts with the generalization phase (phase-0), which aims to learn a global model that performs well on the entire data distribution. The personalization phase (phase-1) is started in asynchronous mode when the loss curve of the generalization phase starts to flatten. Training stops on each local node once local convergence is reached and the test performance metrics are evaluated.

Section~\ref{sec:partition}, ~\ref{sec:sampler}, and ~\ref{sec:personalization} discuss the details of our partitioning, sampling, and personalized training algorithms, respectively.  

\subsection{Entropy-Aware Graph Partitioning}\label{sec:partition}
We aim to create an effective partitioning algorithm that minimizes the total entropy of the partitions. This is achieved by assigning weights to the edges based on two factors: (1) the similarity of their features and (2) the degree of the neighborhood vertex. The weighted partitioning scheme ensures that (1) the nodes that have similar features (and labels) have a higher probability of being in the same partition; this is expected to help in reducing average entropy and improve Micro-F1 scores (2) for nodes having a lower degree, their neighbors have a high probability of being in the same partition, which can help in reducing the communication overhead~\cite{JointSampling} for distributed GNNs and improve scalability.

Consider $u$ and $v$ are any two vertices of the graph, and $|N(v)|$ denotes the size of the neighborhood of $v$. Algorithm~\ref{algo:weighted} shows the edge-weight assignment process. We compute the similarity of two neighborhood vertices as the dot product of their initial features. Next, we compute an approximate probability that a neighborhood vertex $u$ of $v$ is among the K nodes sampled for message aggregation by the GraphSage algorithm~\cite{graphsage,JointSampling}. If $u$ has a low degree, the value of $p$ will be closer to 1. We use a weighted combination of similarity and $p$ as the edge weight, as shown in Line~\ref{line:weighted}. $c$ is a hyper-parameter that depends on the properties of the graph. Finally, we use the weighted-METIS algorithm~\cite{metis}, a  multilevel recursive-bisection-based partitioning technique, to partition the newly constructed weighted graph with roughly the same number of vertices in each partition. 
\vskip -2mm
\begin{algorithm}
    \caption{Edge Weighted Graph Partitioning}
    \label{algo:weighted}
    \SetKwInOut{Input}{Input}
    \SetKwInOut{Output}{Output}
    \Input{Graph $G$}
    \Output{Graph Partitions $(parts)$}
    \For{v $\in$ V} 
    { 
        \For{ $\forall u : (u,v) \in E$} 
        { 
            $similarity = \mathbf{h}_u^0 \cdot \mathbf{h}_v^0$  \label{line:similarity}
           
            $p = 1-e^{\frac{-K}{|N(v)|}}$  \label{line:degree}
            
            $W_{uv}$ = $(c*similarity + p)*100$  \label{line:weighted}
        }
    }
    $parts = WeightedMETIS(G, W)$ \label{line:partition}
    
    Save parts
\end{algorithm}
\vspace{-0.4cm}

The time complexity of the above Algorithm~\ref{algo:weighted} is $O(|E||D|)$, where $|E|$ is the number of edges in the graph as each edge is traversed exactly once, and similarity can be computed in $O(D)$ where $D$ is the feature dimensionality. 

\subsection{CBS: Class Balanced Sampler} \label{sec:sampler}
Our sampler, CBS, is inspired by the pick sampler~\cite{PickNChoose} which assigns the sampling probability of a training node based on its normalized degree and class label frequency. Instead of using all the training nodes in an epoch, CBS generates a random subset using the following probability per training node. 
\begin{equation}\label{eq:sample}
P(v) = \frac{||\hat{A}(:,v)||^2}{CF(class[v])}
\end{equation}
where $\hat{A} = D^{-\frac{1}{2}}AD^{\frac{1}{2}}$ is the normalized adjacency matrix and $D$ is the degree matrix. The size of the subset is a fraction, typically 25\% of the training nodes. A complete training pass on this subset is called a mini-epoch. In every mini-epoch, a new subset is sampled from the entire training set of that compute host. 
Each mini-epoch is further divided into iterations where a random batch is sampled every iteration. This increases the probability that the training nodes of minority classes will participate in every batch. An additional benefit of this approach is that the mini-epochs run much faster and have better convergence, explained in Section~\ref{sec:results}.

\subsection{Personalization under Class Imbalance} \label{sec:personalization}

A global model is trained in the generalization phase (phase-0) till reasonable accuracy is achieved. The models on each compute host are initialized with the same parameters and are kept in sync by applying the same update equation. In the personalization phase (phase-1), the aggregation is stopped, and each node learns independently to tune each local model according to its local training data.
This reduces synchronization overheads and speeds up the training process considerably. A parameter controls the proportion of generalization and personalization in a given number of epochs.

During phase-1, to control model overfitting to local distributions, a regularization loss term has been added to the Loss function in Eq.~\ref{eq:loss} to keep the personal model weights close to the general model learned after phase-0. The squared norm of the difference between these tensors is added to the loss function as a regularization term. Another form of regularization that has been added is early stopping. In phase-0, the early stopping happens based on the average micro-F1 score on the validation set of all partitions; all compute hosts stop simultaneously.
In phase-1, the individual micro-F1 score decides when to stop training. The best model is saved. The early stopping of both phases is done independently of the other phase.

Thus, the final loss function of the $i^{th}$ compute host during phase-1, including the regularization term, is:

\begin{equation} \label{eq:loss}
Loss_{i} = \sum_{j=1}^{n} CrossEntropy_j(\mathbf{W_i}) + \lambda||\mathbf{W}_{i}^{P}-\mathbf{W}^{G}||_{2}
\end{equation}

Here $n$ is the number of training examples in a batch, $\mathbf{W}_{i}^{P}$ are the personal weights of $i$-th computer host, and $\mathbf{W}^{G}$ are the general weights same for all the compute hosts.


%% file: Sections/4_ExperimentalSetup.tex
\input{Tables/datasets}
\section{Experimental Setup}\label{sec:experiments}
We implemented the proposed strategies using DGL 0.9~\cite{DGL}. We use PYMETIS, a python wrapper for METIS, for performing the weighted partitions. We evaluated our approach on several datasets using GraphSAGE~\cite{graphsage}. Table~\ref{tbl:inputs} shows the list of datasets~\cite{DistDGL} used for our evaluation.

We evaluated our algorithms using two platforms. 
For larger datasets, Reddit, OGB-Products, and OGB-Papers, we conducted experiments on a shared commodity cluster using up to 16 compute hosts. For Flickr and Yelp, we use a dedicated Amazon AWS EC2 cloud cluster with 4 VMs.
Each VM is equipped with (R6a.xlarge) Intel(R) Xeon(R) Es-2686 v4 CPU having 8 processing cores with 2.30 GHz frequency, 32GB RAM, and 200 GB external storage. For EW partitioning of larger graphs like Reddit and OGBN-Products VM equipped with (R6a.4xlarge) having 128 GB RAM has been used.

For all our experiments, we choose the learning rate as 0.001, the neighborhood sample as (25,25), and the number of layers as 2, unless specified. Restricting to two layers limits the bottleneck of communication time and speeds up distributed training.
The results reported are an average of over five runs. 
For Flickr, we don't use the sampler as this drastically reduces the number of nodes in an epoch.
For Yelp, the convergence happens slowly with the sampler, and increasing the learning rate makes the curve noisier; hence, the weighted scheme training has been run for more epochs, while in others, even the 100 epochs with sampling are enough.

%% file: Tables/datasets.tex
\begin{table*}[h]
\begin{center}
\begin{tabular}{c|ccccccc}
  \toprule
  \textbf{Data Set} & \textbf{Nodes} & \textbf{Edges} & \textbf{Features} & \textbf{Labels} & \textbf{Train/Val/Test \%} & \textbf{Avg. Degree} & \textbf{Comments} \\
  \midrule
  \textbf{Flickr} & 89,250 & 899,756 & 500 & 7 & 50/25/25 & 20 & Noisy Labels\\
  \textbf{Yelp} & 716,847 & 13,954,819 & 300 & 100 & 75/15/10 & 39 & Multilabel\\
  \textbf{Reddit} & 232,965 & 114,615,892 & 602 & 41 & 66/10/24 & 492 & High Node degree, Feature Dimensions\\
  \textbf{OGBN-Products} & 2,449,029 & 61,859,140 & 100 & 47 & 8/2/90 & 51 & Out of Distribution\\
  \textbf{OGBN-Papers} & 111,059,956 & 1,615,685,872 & 128 & 172 & 78/8/14 & 29 & $\sim$98\% Unlabelled\\
  \bottomrule
\end{tabular}
 \caption{Statistics of Graph Datasets spanning a wide range of applications used in this paper for experimental evaluation. }
\vskip -5mm
\label{tbl:inputs}
\end{center}
\end{table*}

%% file: Sections/5_Results.tex
\section{Results and Discussions}\label{sec:results}
This section describes the experimental analysis of our proposed ideas. We use the following notations to describe the various approaches evaluated in the paper.
\begin{itemize}
    \item \textbf{METIS:} Default METIS graph partitioning scheme used by the DistDGL
    \item \textbf{DistDGL:} Default distributed training followed by DistDGL using METIS partition and Graph Sage 
    \item \textbf{EW:} Edge weighted partitioning scheme
    \item \textbf{CBS:} Class balanced sampling scheme
    \item \textbf{GP:} Two-phase (Generalize and Personalize) distributed training of the model
\end{itemize}



\subsection{Comparing overall performance}
\input{Tables/accuracy}
\input{Tables/figures}

Table \ref{tab:metrics_per_partition} shows the performance comparison of our EW partitioning combined with GP and CBS with the baseline (\textbf{DistDGL}) on 4 compute hosts. We apply the distributed training with generalization and early stopping for the baseline. The table reports micro-F1, weighted-F1 scores, training time, and highlights (in blue) the best-performing score for each dataset and metric. From the results, we observe that our proposed algorithms achieve an average improvement of 4\% on micro-F1 scores and 1.9\% on weighted-F1 scores. This is because \textbf{EW} yields a few partitions with very low entropy and obtains high and moderate scores on the remaining partitions to achieve the best overall accuracy. On OGBN-Papers\footnote{This experiment was repeated twice on 16 compute hosts using METIS partitions in both cases.}, there is a relative improvement of 16.2\% on the micro-F1 score and 11.3\% on the weighted-F1 score by performing \textbf{GP} with class balancing.

Table \ref{tab:metrics_per_partition} also reports the training time for each training technique. It is evident that \textbf{EW+GP+CBS} converges faster than the \textbf{DistDGL} and shows a significant reduction in training time in large graph benchmarks: Reddit by 2.2x, OGBN Products by 3.4x and OGBN Papers by 1.8x). These reductions indicate the usefulness of CBS and the personalization technique. This can be understood further with the help of Figure \ref{tab:graphs} where we plot the validation micro-F1 and training loss for various partitioning algorithms. We trigger the personalization phase when the loss curve starts to flatten. For \textbf{EW+GP+CBS}, Loss includes an extra term due to regularization.

Results show that \textbf{EW+GP+CBS} converges faster than \textbf{DistDGL} and there is a noticeable jump in micro-F1 scores for OGBN-Products and Flickr as soon as we begin the personalization phase. We implement distributed early stopping to minimize the epochs wasted and save the best model obtained. In the case of Yelp, convergence is noisy and slow, so even though there is no jump in the score, we obtain a reduction in training time. Finally, there is a sharper decay in all the loss curves as soon as the personalization starts, which indicates faster convergence. This demonstrates the effectiveness of personalization in adapting to the local distribution of data. 

\subsection{Scalability with \textbf{EW+GP+CBS}}
\input{Tables/scalability_ew_new}
We perform scalability experiments with OGBN Products dataset for 4, 8, and 16 compute hosts and present the results in Table~\ref{tab:scalability-ew}. We notice that the micro-F1 scores start degrading for \textbf{DistDGL} as we increase the number of hosts because of the increased heterogeneity of partitions. In contrast, for \textbf{EW+GP+CBS}, the micro-F1 score is best for the 8 partitions and is always better than \textbf{DistDGL} for all settings. CBS reduces the Epoch time by a factor of 3 as compared to the \textbf{DistDGL}. \textbf{EW+GP+CBS} achieves a consistent 3x reduction in training time for all settings.

\subsection{Comparison with the centralized model}
In Table \ref{tab:compare}, we compare the performance metrics obtained by distributed training of GNN (\textbf{DistDGL and EW+GP+CBS}) with the state-of-the-art centralized training~\cite{En-GCN} of GraphSAGE using 2-layers. The results show that the \textbf{DistDGL} degrades performance by 1-2\% 
\input{Tables/compare}

\subsection{Discussion}
We now present our analysis of the reasons behind the improvements obtained by our techniques and when to use each of them. 
\input{Tables/entropy_data}
Table \ref{tab:perf} shows entropies of various partitioning algorithms. We note that EW consistently reduces the total entropy of partitions as compared to METIS and improves the micro F1 and weighted F1 metric. This is because EW tends to concentrate the nodes with similar features (labels) to a single partition.
Although this may cause a slight degradation in convergence during the generalization phase, it improves the micro-F1 scores during the personalization phase as each model adapts to its own local distribution (Fig~\ref{tab:graphs}). GP also significantly reduces the communication overhead of averaging gradients across all partitions, as shown by the good speed-up achieved on large graph benchmarks. CBS helps alleviate the imbalance in class distribution caused by EW partitioning and helps speed up the convergence significantly (2x-3x) for large graph benchmarks. We also note that GP+CBS can also speed up (1.75x on average) \textbf{DistDGL} with METIS partition while maintaining the same accuracy. We thus conclude that \textbf{EW+GP+CBS} should be used to maximize the micro-F1 score (accuracy) and minimize training time while distributed training of large graph benchmarks.


Table \ref{tab:perf} shows the pre-processing time for different partitioning schemes. \textbf{EW} incurs additional overhead due to the time spent by the PyMETIS library in performing edge-weighted partitioning. In Yelp, about 23.2\% of the time is spent on edge weight assignment, while 76.7\% of the total time is spent on PyMETIS calls. 


%% file: Tables/accuracy.tex

\begin{table}[h]
\centering
\begin{adjustbox}{}
\small
\newcolumntype{C}{>{\centering\arraybackslash}p{1.6cm}}
\newcolumntype{D}{>{\centering\arraybackslash}p{0.6cm}}
\begin{tabular}{Dl|CC}
\toprule
\textbf{} & \textbf{F1 Metric} & \textbf{DistDGL} & \textbf{EW+CBS+GP} \\
\midrule
\multirow{3}{*}{\rotatebox[origin=c]{90}{Flickr}}  & Micro  &     $51.43 \pm 0.18$  & \textcolor{blue} {$51.98 \pm 0.18$}  \\
& Weighted & $43.08\pm 0.36$  & \textcolor{blue} {$43.64 \pm 0.35$}   \\
& Train Time (s) & 477 (1.15$\times$)  & \textcolor{blue} {414}   \\
\midrule
\multirow{3}{*}{\rotatebox[origin=c]{90}{Yelp}}  & Micro & \textcolor{blue}{$58.67 \pm 1.42$}  &   {$58.5 \pm 1.36$}     \\
& Weighted & \textcolor{blue}{$54.88 \pm 1.94$} &  $54.55 \pm 1.8$  \\
& Train Time (s) & 4971(1.35$\times$) & \textcolor{blue} {3670}   \\
\midrule
\multirow{3}{*}{\rotatebox[origin=c]{90}{Reddit}} & Micro & $94.87 \pm 0.11$ & \textcolor{blue} {$96.17 \pm 0.16$}  \\
 & Weighted  &  $95.5 \pm 0.11$  & \textcolor{blue} {$96.07 \pm 0.13$}  \\
 & Train Time (s) & 12552 \nolinebreak(2.2$\times$)  & \textcolor{blue} {5724}   \\
 \midrule
\multirow{3}{*}{\rotatebox[origin=c]{90}{\begin{tabular}[c]{@{}c@{}}OGBN\\ Products\end{tabular}}} & Micro &  $78.36 \pm 0.35$  & \textcolor{blue} {$78.94 \pm 0.32$}  \\
& Weighted & $77.08 \pm 0.42$ & \textcolor{blue} {$77.94 \pm 0.25$}   \\
& Train Time (s) & 7194 (3.4$\times$) & \textcolor{blue} {2115}   \\
\midrule
\multirow{3}{*}{\rotatebox[origin=c]{90}{\begin{tabular}[c]{@{}c@{}} OGBN\\ Papers\end{tabular}}}   & Micro  & $43.9$  &\textcolor{blue} {\textbf{51.02}}   \\
& Weighted        &     $44.23$   &   \textcolor{blue}{\textbf{49.25}}   \\ 
& Train Time (s) &  12763 \nolinebreak(1.8$\times$)  & \textcolor{blue} { 6996}   \\
\bottomrule
\end{tabular}
\end{adjustbox}
\vskip -1mm
\caption{Comparing performance metrics of various algorithms. Scores are reported as percentages and best performing technique is highlighted.}
\vskip -4mm
\label{tab:metrics_per_partition}
\end{table}

%% file: Tables/figures.tex

\begin{figure*}[ht]
\centering
\begin{adjustbox}{}
\begin{tabular}{cccc}
\textbf{Flickr} & \textbf{Yelp} & \textbf{Reddit} & \textbf{OGBN-Products}  \\                                                            
\includegraphics[scale=0.31]{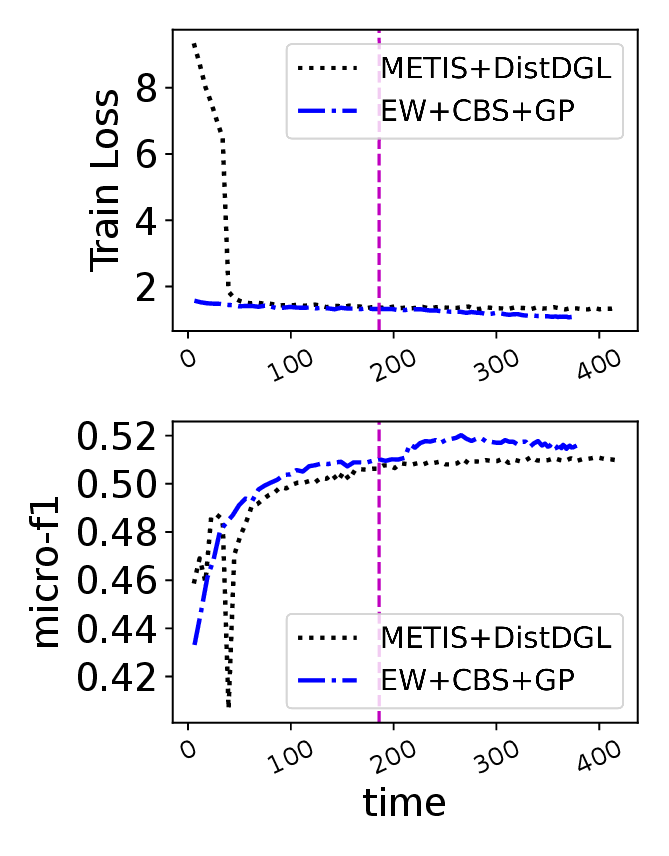} &  \includegraphics[scale=0.31]{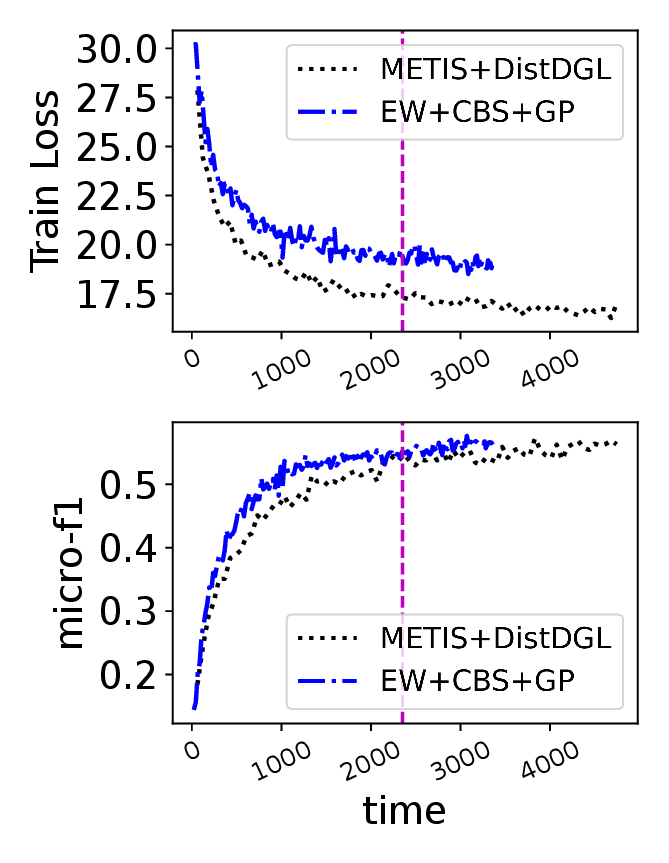} & \includegraphics[scale=0.31]{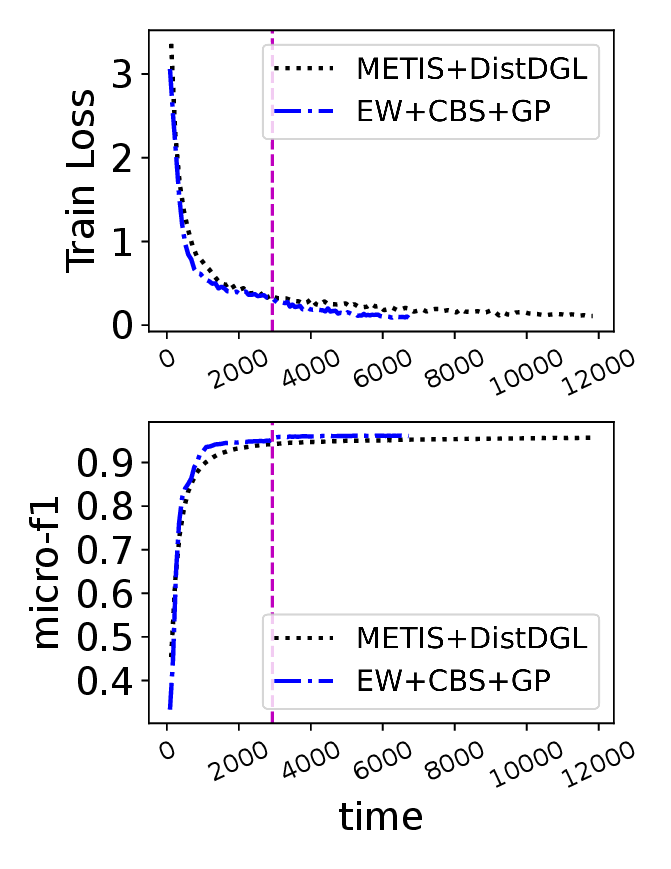} & \includegraphics[scale=0.31]{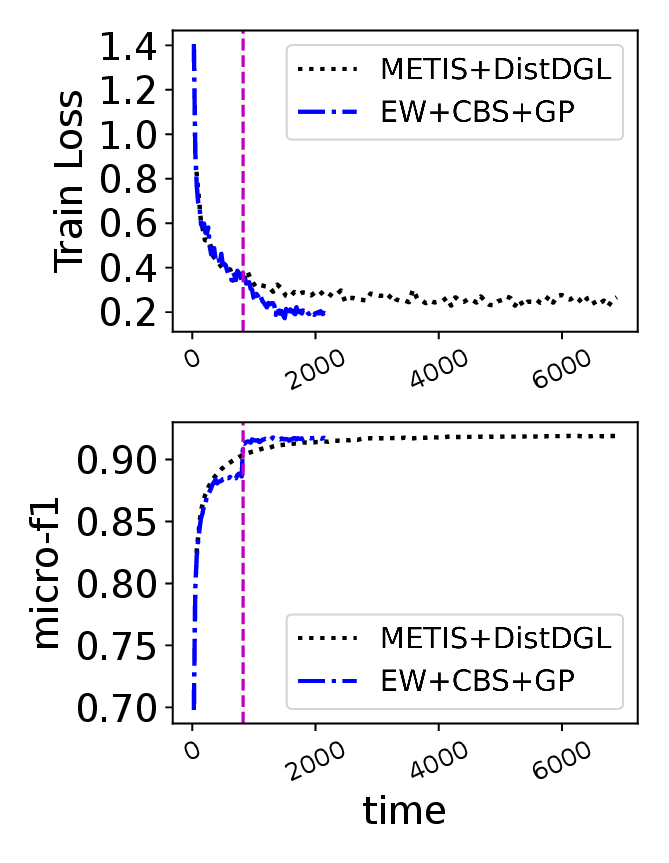}\\                  
\end{tabular}
\end{adjustbox}
\vspace{-3mm}
\caption{The convergence curves for training loss, validation micro scores for Flickr, Reddit, Yelp and OGBN-Products using various partitioning schemes. The magenta vertical line in all plots represents the time where personalization starts.}
\vspace{-3mm}
\label{tab:graphs}
\end{figure*}

%% file: Tables/scalability_ew_new.tex

\begin{table}[h]
\centering
\begin{adjustbox}{}
\small
\newcolumntype{C}{>{\centering\arraybackslash}p{1.6cm}}
\newcolumntype{D}{>{\centering\arraybackslash}p{1.0cm}}
\begin{tabular}{l|C|p{0.1\textwidth}|CC}
\toprule
\textbf{Metric} & \textbf{Partitions} & \textbf{DistDGL} & {\textbf{EW+CBS+GP }}\\
\midrule
\multirow{3}{*}{\rotatebox[origin=c]{}{\begin{tabular}[c]{@{}c@{}}Training\\ Time (s)\end{tabular}}}  & 4 & 7194 (3.8$\times$) &  \textcolor{blue} {1876}   \\                    
                                                 & 8    &   3632 (3.2$\times$)    & \textcolor{blue} {1113}  \\    
                                                  & 16   &   2261 (2.8$\times$)   & \textcolor{blue} {801}  \\

\midrule

\multirow{3}{*}{\rotatebox[origin=c]{}{\begin{tabular}[c]{@{}c@{}}Epoch\\ Time (s)\end{tabular}}}  & 4 & 37.84 &  \textcolor{blue} {12.54} \\ 
                                                  & 8    &   18.92    &  \textcolor{blue} {6.76}  \\
                                                   & 16    &   11.55   &  \textcolor{blue} {4.06}  \\

\midrule
\multirow{3}{*}{\rotatebox[origin=c]{}{\begin{tabular}[c]{@{}c@{}}Micro\\ F1\end{tabular}}} & 4     &   78.36    &  \textcolor{blue} {79.48}     \\ 
                                                    & 8     &   77.2   & \textcolor{blue} {79.71}   \\    
                                                    & 16     &   75.3   & \textcolor{blue} {76.4}    \\ 
                                                
\bottomrule
\end{tabular}
\end{adjustbox}
\caption{Scaling results for OGBN Products}
\vskip -3mm
\label{tab:scalability-ew}
\end{table}

%% file: Tables/compare.tex
\begin{table}[H]
\fontsize{9}{10}\selectfont
 \centering
\begin{tabular}{l|cccc}
\toprule
\textbf{Method} & \textbf{Flickr} & \textbf{Reddit} & \textbf{OGBN-Products}  \\ 
\midrule
Centralized    &  \color{blue} \textbf{52.26}  &  \color{blue} \textbf{96.34} &   78.20       \\
DistDGL    &   51.43    &   94.47 &   78.36 \\
EW+GP+CBS          &   51.98    &   96.17    &   \color{blue} \textbf{79.71}       \\

\bottomrule
\end{tabular}
\caption{Comparison with centralized GraphSAGE model}
\vskip -2mm
\label{tab:compare}
\end{table}

%% file: Tables/entropy_data.tex
\newcolumntype{C}{>{\centering\arraybackslash}p{0.8cm}}
\newcolumntype{D}{>{\centering\arraybackslash}p{1.5cm}}
\newcolumntype{E}{>{\centering\arraybackslash}p{1cm}}

\begin{table}[ht]
    \fontsize{9}{10}\selectfont
    \setlength{\tabcolsep}{2.7pt}
    \centering
    \begin{tabular}{D|CC|CC|CC}
        \toprule
        \multirow{2}{*} 
        \textbf{\textbf{Partition}} & \multicolumn{2}{c}{\textbf{Reddit}} & \multicolumn{2}{c}{\textbf{Yelp}} & \multicolumn{2}{c}{\textbf{OGB-Products}} \\
        \textbf{Method} & $\mathbf{H}(\mathcal{P})$  & \textbf{time} & $\mathbf{H}(\mathcal{P})$  & \textbf{time} & $\mathbf{H}(\mathcal{P})$  & \textbf{time}\\  \midrule
        \textbf{METIS}            & 3.35           &   31.94      & 38.05          &  4.8     & 2.80            &   42.1         \\
        \textbf{EW}            &  \textcolor{blue}{3.21}          &   172.9     & \textcolor{blue}{37.92}        &   91.2    & \textcolor{blue}{2.44}           &   784.7     \\
        \bottomrule
    \end{tabular}
    \caption{Average entropy and time to partition (in sec), across various graph partitioning algorithms.\\
   }\label{tab:perf}
    \vskip -5mm
\end{table}

%% file: Sections/6_Conclusion.tex
\section{Conclusion}\label{sec:conclusion}
This paper presents the importance of entropy-aware training algorithms to improve the performance of distributed GNN. The edge-weighted partitioning technique proves to be effective for improving Micro-F1. The training techniques: personalization and class-balanced sampling help simultaneously improve performance metrics and reduce training time. 
Through extensive experimental evaluation on the DistDGL framework,  
we achieved a (2-3x) speedup in training time and 4\% improvement on average in micro-F1 scores of 5 large graph benchmarks compared to the standard baselines. 

%% file: Sections/7_Appendix.tex
This artifact describes the key contributions of the paper followed by the required steps to reproduce the experimental results presented in the paper. 

\section{Artifact Identification}
\subsection{Abstract}
Graph Neural Networks (GNNs) are powerful models for learning over graphs. To speed up training on very large, real-world graphs (billion scale edges) several distributed frameworks have been developed. A fundamental step in every distributed GNN framework is graph partitioning. On several benchmarks, we observe that these partitions have heterogeneous data distributions which affect model convergence and performance. They also suffer from class imbalance and out-of-distribution problems, resulting in lower performance than centralized implementations.

We holistically address these challenges, by developing entropy-aware partitioning algorithms that minimize total entropy and/or balance the entropies of graph partitions. We observe that by minimizing the average entropy of the partitions, the micro average F1 score (accuracy) can be improved. Similarly, by minimizing the variance of the entropies of the partitions and implementing a class-balanced sampler with Focal Loss, the macro average F1 score can be improved. We divide the training into a synchronous, model generalization phase, followed by an asynchronous, personalization phase that adapts each compute host's models to their local data distributions. This boosts all performance metrics and also speeds up the training process significantly.

We have implemented our algorithms on the DistDGL framework where we achieved a 4\% improvement on average in micro-F1 scores and 11.6\% improvement on average in the macro-F1 scores of 5 large graph benchmarks compared to the standard baselines.

\subsection{System and Data Model}

To evaluate our system, we need to partition the given input graph (using DGL graph data format), and subsequently perform distributed training on multiple compute hosts. To partition a graph into $N$ partitions, we require a single compute host. However, to perform distributed training, we require cluster of $N$ compute hosts. For each compute host, we assign a partitioned graph along with a set of train, validation and test nodes. We distributed these partitions to the compute hosts via a network file system (NFS). For distributed training on $N$ compute hosts, we configure one compute host as a server and the remaining hosts as clients. All the clients access the common storage via NFS. 


\section{Artifact Dependencies and Requirements}

\subsubsection{Hardware Configuration}
To evaluate the artifcats on the smaller datasets, we require 4 compute hosts with 32 GB memory and NFS.  To evaluate them on larger graphs (i.e., OGB-Papers), we require 16 machines with 128GB RAM  with NFS. For smaller datasets we performed our experiments on AWS using 32GB RAM r6a.xlarge machine using 4 instances. We configured the instances with NFS. For larger datasets we have used a shared commodity cluster, using up to 128 GB memory on each compute host.


Note that for ease of artifact evaluation, we can provide the AWS container with the required environment for smaller datasets. However, our system can also configured by provisioning and setting up the AWS instances with the following instructions.

\subsection{Server Configuration:} Create an Ec2 instance with the following settings.
\begin{verbatim}
name: pc_node
os: ubuntu 22.04
instance type: r6a.xlarge
default vpc
subnet: ap-south 1c
existing security grp: launch-wizard-1

In security grp select: nfs-server 
Instance type: r6a.4xlarge
\end{verbatim}

While creating the instance, it is required to enter the following script in the textbox corresponding to the additional details. These instructions will be execute at the time of launching the machine.

\begin{verbatim}
Common for server and client 
#!/bin/bash
//Assuming home directory as /home/ubuntu
cd /home/ubuntu 
export HOME=/home/ubuntu
wget \
https://repo.continuum.io/miniconda/\
Miniconda3-latest-Linux-x86_64.sh \
-O /home/ubuntu/miniconda.sh
chmod +x miniconda.sh
bash miniconda.sh -b \
-p /home/ubuntu/miniconda3
source miniconda3/bin/activate
conda create -y -n envforgnn python=3.9
conda activate envforgnn
conda install -y pytorch==1.9.0 \
torchvision==0.10.0 \
torchaudio==0.9.0 \
cpuonly -c pytorch
conda install -y -c dglteam dgl=0.9
conda install -y pandas
conda install -y scikit-learn
conda install -y matplotlib-base
conda install -c conda-forge ogb
export DGLBACKEND=pytorch

sudo apt-get install -y nfs-kernel-server
mkdir -p /home/ubuntu/workspace
sudo -- bash -c 'echo \
"/home/ubuntu/workspace \
172.31.16./20 \
(rw,no_root_squash,sync,no_subtree_check)" \
>> /etc/exports'
sudo systemctl restart nfs-kernel-server
\end{verbatim}

Once the server is configured, it required to be launched before the clients can be configured. \\

\subsection{Client Configuration} Create 3 Ec2 instances each with the following configuration.

\begin{verbatim}
name: pc_node
os: ubuntu 22.04
instance type: r6a.xlarge
default vpc
subnet: ap-south 1c
existing Security grp: launch-wizard-1
Storage: 8 gb ssd 
\end{verbatim}

Similar to server configuration, it is required to enter the following script in the textbox corresponding to the additional details. These instructions will be execute at the time of launching the machine.

\begin{verbatim}
    
#!/bin/bash
cd /home/ubuntu
export HOME=/home/ubuntu
wget \
https://repo.continuum.io/miniconda/\
Miniconda3-latest-Linux-x86_64.sh \
-O /home/ubuntu/miniconda.sh
chmod +x miniconda.sh
bash miniconda.sh -b \
-p /home/ubuntu/miniconda3
source miniconda3/bin/activate
conda create -y -n envforgnn python=3.9
conda activate envforgnn
conda install -y pytorch==1.9.0 \
torchvision==0.10.0 \
torchaudio==0.9.0 \
cpuonly -c pytorch
conda install -y -c dglteam dgl=0.9
conda install -y pandas
conda install -y scikit-learn
conda install -y matplotlib-base
conda install -c conda-forge ogb
export DGLBACKEND=pytorch

sudo apt-get install -y nfs-common
mkdir -p /home/ubuntu/workspace
sudo mount -t nfs \
<put nfs server private ipv4>:\
/home/ubuntu/workspace \
/home/ubuntu/workspace
mount -a
\end{verbatim}

Once the server and the clients are launched, they need to be configured with passwordless authentication using the instructions given at \url{https://linuxize.com/post/how-to-setup-passwordless-ssh-login/} 



The Data Model consists of mainly the graphs. The graphs have been stored in the DGL graph data format. In this format, a folder is made for a set of partitions of a graph. There is JSON file detailing the node and edge splits for each partition, and then there are respective folders for each partition which store node and edge features in binary format.

\section{Artifact Installation}

The artifacts can be downloaded from the following git hub repository and can be set up using the instructions provided in the \emph{ReadME}. 
\url{https://github.com/Anirban600/EAT-DistGNN}
The repository contains the source code for partitioning and performing distributed training. The partitioning code contains two partitioning strategies described in the paper, i.e.,  METIS, Edge Weighted (EW). 
The training code consists of all our distributed training algorithms: 1) Class Balanced Sampler (CBS), 2) Generalized-Personalized (GP) model, along with scripts to run them with required hyper-parameters.

Through the artifacts, we expect to reproduce all the major results listed in the paper, however, a statistical variability can be expected. Moreover, graph partitioning and training time will depend on the machine configurations.


\section{Reproducibility of Experiments}



The experimental workflow has the following steps: (1) graph partitioning, (2) distributed training, and (3) post processing to generate the plot graphs and tables listed in the paper. To ease the workflow for evaluators we automated all the above steps for each graph. The \emph{README} present in the repository shows the instructions to execute the automated script on each dataset. 



The times required for completing all experiments listed in the paper, for each of the input graph are given below. Note that these are the times are upper bounds based on slowest hardware configuration used by us for experiments and may vary depending on hardware.
\begin{enumerate}
    \item Flickr - 1 hour
    \item Yelp - 4 hours
    \item OGBN-Products - 6 hours
    \item Reddit - 5 hours
    \item OGB-Papers - 20 hours
\end{enumerate}


Running the automated script file for each respected graph will generate the following tables and figures reported in the paper.
\begin{enumerate}
    \item Table 5 for average entropy and time to partition(in sec, across various graph partitioning algorithms.
    \item Table 2 for comparing performance metrics of various algorithms for different graph datasets.
    \item Figure 3 for the convergence curves for training loss, validation micro score for Flickr, Reddit, Yelp and OGBN-Products using various partitioning schemes
\end{enumerate}
